%% file: paper568.tex
\documentclass[runningheads]{llncs}
\usepackage{graphicx}
\usepackage{epsfig}
\usepackage{graphicx}
\usepackage{multirow}
\usepackage{amsmath}
\usepackage{amssymb}
\usepackage{booktabs}
\usepackage{ dsfont }
\usepackage{ cite }
\usepackage{float}
\usepackage{color}
\usepackage{xcolor}
\usepackage{soul}
\usepackage{float}
\usepackage{url}
\usepackage{rotate}
\usepackage{tikz}
\usepackage[colorlinks=true, citecolor=blue]{hyperref}%
\usepackage{makecell}
\usepackage{multirow}
\usepackage{tabularx}

\usepackage{caption}
\usepackage{subcaption}

\DeclareMathOperator*{\argmin}{arg\,min}

\usepackage[normalem]{ulem}

\definecolor{alizarin}{rgb}{0.82, 0.1, 0.26}
\definecolor{aliceblue}{rgb}{0.94, 0.97, 1.0}
\definecolor{amber}{rgb}{1.0, 0.75, 0.0}
\definecolor{amethyst}{rgb}{0.6, 0.4, 0.8}

\newcommand{\red}[1]{{\color{red}#1}}
\newcommand{\blue}[1]{{\color{blue}#1}}

\newcolumntype{P}[1]{>{\centering\arraybackslash}p{#1}}

\usepackage{wrapfig,lipsum}

\usepackage[misc]{ifsym}

\begin{document}
\title{
Multi-view Local Co-occurrence and Global Consistency Learning Improve Mammogram Classification Generalisation
}

%
%
\author{
Yuanhong Chen\textsuperscript{1(\Letter)}, 
Hu Wang\textsuperscript{1}, 
Chong Wang\textsuperscript{1}, 
Yu Tian\textsuperscript{1}, 
Fengbei Liu\textsuperscript{1}, 
Yuyuan Liu\textsuperscript{1},
Michael Elliott\textsuperscript{2}, 
Davis J. McCarthy\textsuperscript{2}, 
Helen Frazer\textsuperscript{2}\textsuperscript{3},\\
Gustavo Carneiro\textsuperscript{1}
}

\titlerunning{BRAIxMVCCL for Multi-view Mammogram Classification}

\authorrunning{Y. Chen et al.}

\institute{
 \textsuperscript{1} Australian Institute for Machine Learning, The University of Adelaide, Adelaide, Australia \\
 \{yuanhong.chen, hu.wang, chong.wang, yu.tian, fengbei.liu, yuyuan.liu, gustavo.carneiro\}@adelaide.edu.au\\
 \textsuperscript{2} St Vincent's Institute of Medical Research, Melbourne, Australia \\
 \{melliott, dmccarthy\}@svi.edu.au\\
 \textsuperscript{3} St Vincent's Hospital Melbourne, Melbourne, Australia\\
 helen.frazer@svha.org.au
}
\maketitle              
\begin{abstract}

When analysing screening mammograms, radiologists can naturally process information across two ipsilateral views of each breast, namely the cranio-caudal (CC) and mediolateral-oblique (MLO) views.
These multiple related images provide complementary diagnostic information and can improve the radiologist's classification accuracy. Unfortunately, most existing deep learning systems, trained with globally-labelled images, lack the ability to jointly analyse and integrate global and local information from these multiple views. By ignoring the potentially valuable information present in multiple images of a screening episode, one limits the potential accuracy of these systems.

Here, we propose a new multi-view global-local analysis method that mimics the radiologist's reading procedure, based on a global consistency learning and local co-occurrence learning of ipsilateral views in mammograms. Extensive experiments show that our model outperforms competing methods, in terms of classification accuracy and generalisation, on a large-scale private dataset and two publicly available datasets, where models are exclusively trained and tested with global labels.

\keywords{Deep learning \and Supervised learning \and Mammogram classification \and Multi-view}
\end{abstract}

\section{Introduction}
Breast cancer is the most common cancer worldwide and the fifth leading cause of cancer related death~\cite{sung2021global}.
Early detection of breast cancer saves lives and population screening programs demonstrate reductions in mortality~\cite{lauby2015breast}.
One of the best ways of improving the chances of survival from breast cancer is based on its early detection from screening mammogram exams~\cite{selvi2014breast}. 
A screening mammogram exam contains two ipsilateral views of each breast, namely bilateral craniocaudal (CC) and mediolateral oblique (MLO), 
where radiologists analyse both views in an integrated manner by searching for global architectural distortions and local masses and calcification.
Furthermore, radiologists tend to be fairly adaptable to images from different machines and new domains with varying cancer prevalence.
Some automated systems have been developed~\cite{frazer2021evaluation,carneiro2017automated,yang2021momminetv2,shen2019deep,shen2021interpretable} but none have achieved accuracy or generalisability in low cancer prevalent screening populations that is 
at the same level as
radiologists~\cite{freeman2021use}. 
We argue that for systems to reach the accuracy and generalisability of radiologists, they will need to mimic the manual analysis explained above.

Global-local analysis methods combine the analysis of the whole image 
and local patches to jointly perform the classification. 
Shen et al.~\cite{shen2019globally, shen2021interpretable} propose a classifier that relies on concatenated features by a global network (using the whole image) and a local network (using image patches).
Another global-local approach shows state-of-the-art (SOTA) 
classification accuracy with a two-stage training pipeline that first trains a patch classifier, which is then fine-tuned to work with a whole image classification~\cite{shen2019deep}. 
Although these two models~\cite{shen2019globally, shen2019deep} show remarkable performance, they do not explore cross-view information and depend on relatively inaccurate region proposals, which can lead to sub-optimal solutions. 
To exploit cross-view mammograms, previous approaches process the two views without trying to find the inter-relationship between lesions from both views~\cite{carneiro2017automated}.

Alternatively, Ma et al.~\cite{ma2021cross} implement a relation network~\cite{hu2018relation} 
to learn the inter-relationships between the region proposals, which can be inaccurate and introduce noise that may lead to a poor learning process, and the lack of global-local analysis may lead to sub-optimal performance.
MommiNet-v2~\cite{yang2021momminetv2} is a system focused on the detection and classification of masses from mammograms that explores cross-view information from ipsilateral and bilateral mammographic views, but its focus on the detection and classification of masses limits its performance on global classification that also considers other types of lesions, such as architectural distortions and calcification.

The generalisation of mammogram classification methods is a topic that has received little attention, with few papers showing methods that can be trained in one dataset and tested in another one. A recent paper exposes this issue and presents a meta-repository of mammogram classification models,
comparing five models on seven datasets~\cite{stadnick2021metarepository}. Importantly, these models are not fine-tuned to any of the datasets.
Testing results based on area under the receiver operating characteristic curve (AUC) show that the GMIC model~\cite{shen2021interpretable} is about 20\% better than Faster R-CNN~\cite{ribli2018detecting} on the NYU reader study test set~\cite{wu2019dataset}, but Faster R-CNN outperforms GMIC on the INBreast dataset~\cite{moreira2012inbreast} by 4\%, and  GLAM~\cite{liu2021weakly} reaches 85.3\% on the NYU test set but only gets 61.2\% on INBreast and 78.5\% on CMMD~\cite{web:cmmd}. Such experiments show the importance of assessing the ability of models to generalise to testing sets from different populations and with images produced by different machines, compared with the training set.

In this paper, we introduce the \underline{M}ulti-\underline{V}iew local \underline{C}o-occurrence and global \underline{C}onsistency \underline{L}earning (BRAIxMVCCL) model which is trained in an end-to-end manner for a joint global and local analysis of ipsilateral mammographic views.
BRAIxMVCCL has a novel global consistency module that is trained to penalise differences in feature representation from the two views, and also to form a global cross-view representation.
Moreover, we propose a new local co-occurrence learning method to produce a representation based on the estimation of the relationship between local regions from the two views. 
The final classifier merges the global and local representations to output a prediction.
To summarise, the \textbf{key contributions of the proposed BRAIxMVCCL} for multi-view mammogram classification are: 1) a novel global consistency module that produces a consistent cross-view global feature, 2) a new local co-occurrence module that automatically estimates the relationship between local regions from the two views, and 3) a classifier that relies on global and local features from the CC and MLO mammographic views.
We evaluate BRAIxMVCCL on a large-scale (around 2 million images) private dataset and on the publicly available INBreast~\cite{moreira2012inbreast} and CMMD~\cite{web:cmmd} datasets without model fine-tuning to test for generalisability.
In all problems, we only use the global labels available for the training and testing images.
Our method
outperforms the SOTA on all datasets in classification accuracy and generalisability.

\section{Proposed Method}
    \begin{figure}[t]
        \centering
        \includegraphics[width=0.85\linewidth]{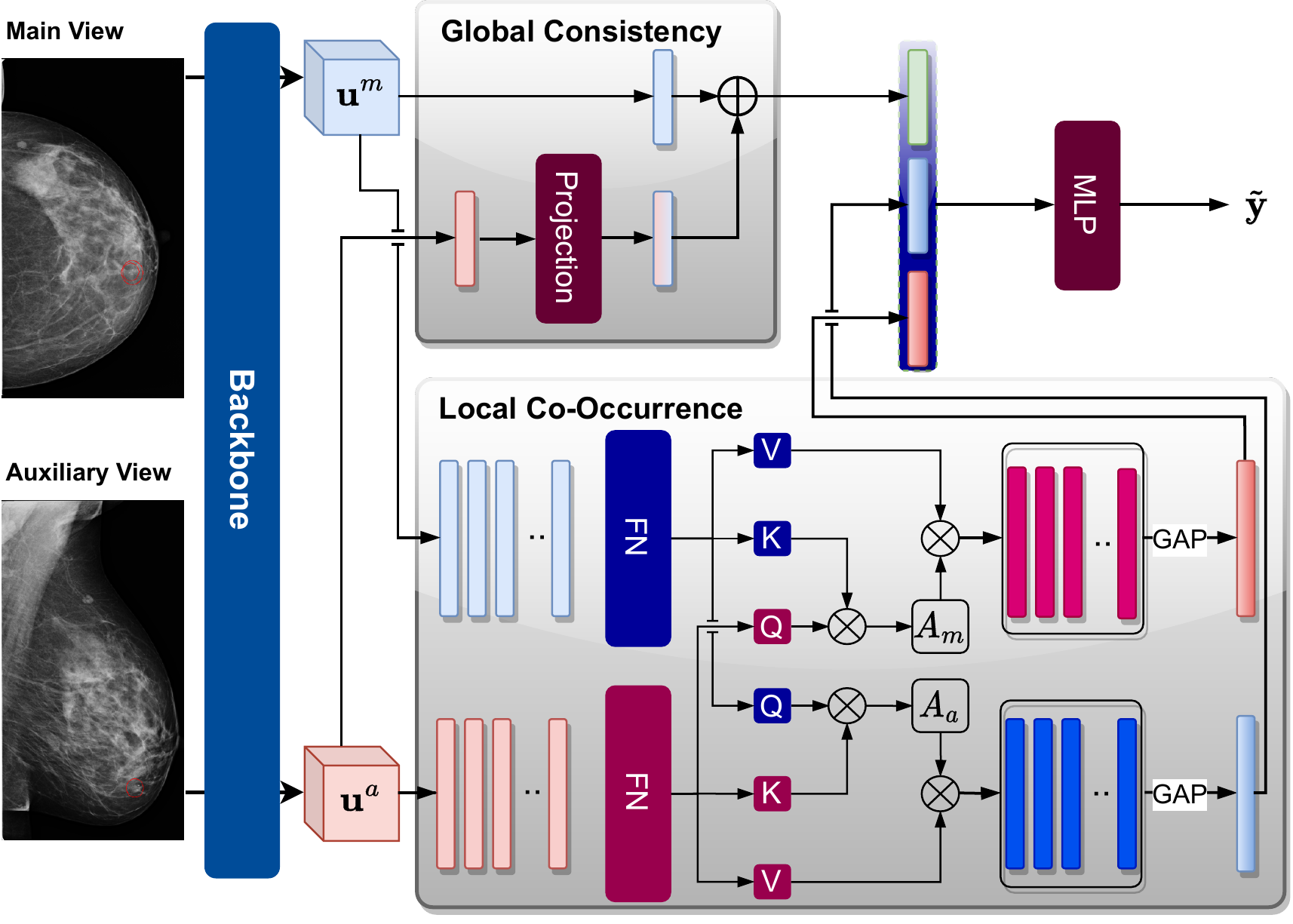}
        \caption{\textbf{BRAIxMVCCL} takes two mammographic views (main and auxiliary) and uses a backbone model to extract the main and auxiliary features $\mathbf{u}^m$ and $\mathbf{u}^a$, where the main components are: 1) a global consistency module that learns a  projection from the auxiliary to the main view, and then combines the main and projected auxiliary features to produce a global representation; 2) a local co-occurrence module that models the local semantic relationships between the two views to produce local representations; and 3) a fusion of the local and global representations to output the prediction $\mathbf{\tilde{y}}$. GAP stands for global average pooling and MLP for multi-layer perception.}
        \label{fig:framework}
    \end{figure}
    

    We assume the availability of a multi-view mammogram training set that contains weakly labelled ipsilateral views (i.e., CC and MLO) of each breast, denoted by $\mathcal{D}=\{\mathbf{x}_i^{m}, \mathbf{x}_i^{a}, \mathbf{y}_i\}_{i=1}^{|\mathcal{D}|}$, where $\mathbf{x}\in\mathcal{X}\subset \mathbb{R}^{H \times W}$ represents a mammogram of height $H$ and width $W$, $\mathbf{x}_i^m$ represents the main view, $\mathbf{x}_i^a$ is the auxiliary view ($m, a \in \{CC, MLO\}$, interchangeably), and $\mathbf{y}_i \in \mathcal{Y} = \{0, 1\}$ denotes the label (1 = cancer and 0 = non-cancer).  The testing set is similarly defined.
     
    \subsection{Multi-View local Co-occurrence and global Consistency Learning}

    The proposed BRAIxMVCCL model, depicted in Fig.~\ref{fig:framework}, consists of a backbone feature extractor denoted as $\mathbf{u}^m=f^B_{\phi}(\mathbf{x}^{m})$ and $\mathbf{u}^a=f^B_{\phi}(\mathbf{x}^{a})$,
    a global consistency module $\mathbf{z}^G=f^G_{\eta}(\mathbf{u}^m,\mathbf{u}^a)$ that forms the global representation from the two views, a local co-occurrence module $\mathbf{z}^m,\mathbf{z}^a=f^L_{\gamma}(\mathbf{u}^m,\mathbf{u}^a)$ that explores the cross-view feature relationships at local regions, and the cancer prediction classifier $\tilde{\mathbf{y}}=p_\theta(\mathbf{y}=1|\mathbf{x}^{m}, \mathbf{x}^{a})=f_{\psi}(\mathbf{z}^G,\mathbf{z}^m,\mathbf{z}^a) \in [0,1]$, where $p_\theta(\mathbf{y}=0|\mathbf{x}^{m}, \mathbf{x}^{a}) = 1-p_\theta(\mathbf{y}=1|\mathbf{x}^{m}, \mathbf{x}^{a})$. 
    The BRAIxMVCCL parameter $\theta = \{\phi,\eta,\gamma,\psi\} \in \Theta$ represents all module parameters and is estimated with the binary cross entropy loss (BCE) and a global consistency loss, as follows 
    \begin{equation}
        \theta^{*}=\argmin_{\theta} \frac{1}{\mathcal{D}}\sum_{(\mathbf{x}_i^{m}, \mathbf{x}_i^{a}, \mathbf{y}_i)\in\mathcal{D}} 
        \ell_{bce}(\mathbf{y}_i,\mathbf{x}^m_i,\mathbf{x}^a_i) + \ell_{sim}(\mathbf{x}^m_i,\mathbf{x}^a_i),       
    \end{equation}
    where $\ell_{bce}(\mathbf{y}_i,\mathbf{x}^m_i,\mathbf{x}^a_i) = -\big(\mathbf{y}_i \log p_\theta(\mathbf{y}_i|\mathbf{x}_i^{m}, \mathbf{x}_i^{a}) + (1-\mathbf{y}_i) \log (1-p_\theta(\mathbf{y}_i|\mathbf{x}_i^{m}, \mathbf{x}_i^{a})\big)$ and $\ell_{sim}(.)$ denotes the global consistency loss defined  in~\eqref{eq:ell_sim}.
    Below, we explain how we train each component of BRAIxMVCCL.

    When analysing mammograms, radiologists focus on global features (e.g., architectural distortions), local features (e.g., masses and calcification), where this analysis depends on complementary information present in both views~\cite{hackshaw2000investigation}. 
    To emulate the multi-view global analysis, our \textbf{global consistency module (GCM)} is trained to achieve two objectives: 1) the global representations of both mammographic views should be  similar to each other, and 2) the complementarity of both views has to be effectively characterised.
    Taking the backbone features $\mathbf{u}^m,\mathbf{u}^a \in \mathbb{R}^{\widehat{H} \times \widehat{W} \times D}$, where $\widehat{H} < H$ and $\widehat{W} < W$, we first compute the features $\mathbf{g}^m,\mathbf{g}^a \in \mathbb{R}^{D}$ with global max pooling.  To penalise differences between the two global representations, we define the two mapping functions
    $\tilde{\mathbf{g}}^{m}=f^{a \to m}(\mathbf{g}^a)$ and 
    $\tilde{\mathbf{g}}^{a}=f^{m \to a}(\mathbf{g}^m)$ 
    that can transform one view feature into another view feature. 
    We maximise the similarity between features 
    for training the global consistency module, as follows: 
    \begin{equation}
        \ell_{sim}(\mathbf{x}^m_i,\mathbf{x}^a_i) =  -0.5 (c(\tilde{\mathbf{g}}_i^{m}, \mathbf{g}_i^m) + c(\tilde{\mathbf{g}}_i^{a}, \mathbf{g}_i^a)),
        \label{eq:ell_sim}
    \end{equation}
    where 
    $c(\mathbf{x}_1,\mathbf{x}_2)  = \frac{\mathbf{x}_1 \cdot \mathbf{x}_2}{\max(\left\lVert \mathbf{x}_1 \right\rVert_2 \cdot \left\lVert \mathbf{x}_2 \right\rVert_2, \epsilon)}$ with $\epsilon>0$ being a factor to avoid division by zero. 
    The output from the global consistency module is formed with a skip connection~\cite{he2016deep}, as in $\mathbf{z}^{G} = \mathbf{g}^m+\tilde{\mathbf{g}}^{m}$ to not only improve model training, but also to characterise the complementary of both views.

    The \textbf{local co-occurrence module (LCM)} aims to build feature vectors from the two views by analysing long-range interactions between samples from both views. Starting from the backbone features  $\mathbf{u}^m$ and $\mathbf{u}^a$, we transform $\mathbf{u}^m$ into the main local feature matrix $\mathbf{U}^m = [\mathbf{u}^m_{1}, \mathbf{u}^m_{2}, ... ,\mathbf{u}^m_{D}]$, with $\mathbf{u}^m_{d} \in \mathbb{R}^{\widehat{H}\widehat{W}}$ 
    and similarly for the auxiliary local feature matrix $\mathbf{U}^a$, where $\mathbf{U}^m,\mathbf{U}^a \in \mathbb{R}^{\widehat{H}\widehat{W} \times D}$. 
    To estimate the local co-occurrence of lesions  between the mammographic views, we propose the formulation of a cross-view feature representation with:
    \begin{equation}
        \widetilde{\mathbf{U}}^m = f^{A}(\mathbf{U}^m,\mathbf{U}^a,\mathbf{U}^a) 
        \text{, and }
        \widetilde{\mathbf{U}}^a = f^{A}(\mathbf{U}^a,\mathbf{U}^m,\mathbf{U}^m),
    \end{equation}
    where $\widetilde{\mathbf{U}}^m ,\widetilde{\mathbf{U}}^a \in \mathbb{R}^{\widehat{H}\widehat{W} \times D'}$,
    $f^{A}(\mathbf{Q},\mathbf{K},\mathbf{V})  = f^{mlp}\Big( f^{mha}(\mathbf{Q},\mathbf{K},\mathbf{V})\Big)$, with
    $f^{mlp}(.)$ representing a multi-layer perceptron and the multi-head attention (MHA)~\cite{vaswani2017attention,dosovitskiy2020image} is defined as
    \begin{equation}
        f^{mha}(\mathbf{Q},\mathbf{K},\mathbf{V}) = softmax\left(\frac{(\mathbf{Q}\mathbf{W}_q)(\mathbf{K}\mathbf{W}_k^{T})}{\sqrt{D}})(\mathbf{V}\mathbf{W}_v)\right),
    \end{equation}
    where $\mathbf{W}_q, \mathbf{W}_k, \mathbf{W}_v \in \mathbb{R}^{D \times D'}$ 
    are the linear layers.

    The view-wise local features for each view is 
    obtained with $\mathbf{z}^m = GAP(\widetilde{\mathbf{U}}^m)$ (similarly for $\mathbf{z}^a$) 
    , where $\mathbf{z}^m,\mathbf{z}^a \in \mathbf{R}^{D'}$, and
    GAP represents the global average pooling.
    Finally, the integration of the global and local modules is obtained via a concatenation fusion, where we concatenate $\mathbf{z}^G, \mathbf{z}^m, \mathbf{z}^a$ before applying  the last MLP classification layer, denoted by $f_{\psi}(\mathbf{z}^G,\mathbf{z}^m,\mathbf{z}^a)$, to estimate $p_{\theta}(\mathbf{y}|\mathbf{x}^m,\mathbf{x}^a)$.

\section{Experimental Results}
    \subsection{Datasets}
    \textbf{Private Mammogram Dataset}\label{Admani}
    The ADMANI (Annotated Digital Mammograms and Associated Non-Image
    data) dataset is collected from several breast screening clinics 
    from 2013 to 2017 and is composed of 
    ADMANI-1 and ADMANI-2. ADMANI-1 contains 139,034 exams (taken from 2013 to 2015) with 5,901 cancer cases (containing malignant findings) and 133,133 non-cancer cases (with benign lesions or no findings). This dataset is split 80/10/10 for train/validation/test in a patient-wise manner. ADMANI-2 contains 1,691,654 cases with 5,232 cancer cases and 1,686,422 non-cancer cases and is a real world, low cancer prevalence screening population cohort.
    Each exam on ADMANI-1,2 has two views (CC and MLO) per breast produced by one of the following manufactures: SIEMENS, HOLOGIC, FUJIFILM Corporation, Philips Digital Mammography Sweden AB, KONICA MINOLTA, GE MEDICAL SYSTEMS, Philips Medical Systems and Agfa.
    
    \textbf{Public Mammogram Datasets} The Chinese Mammography Database (CMMD)~\cite{web:cmmd} 
    contains 5,202 mammograms from 1,775 patients with normal, benign or malignant biopsy-confirmed tumours\footnote{Similarly to the meta-repository in~\cite{stadnick2021metarepository}, we remove the study \#D1-0951 as the pre-processing failed in this examination.}. 
    Following~\cite{stadnick2021metarepository}, we use the entire CMMD dataset for testing, which contains 2,632 mammograms with malignant findings and 2,568 non-malignant mammograms.
    All images were produced by GE Senographe DS mammography system~\cite{web:cmmd}.
    The \textit{INBreast} dataset~\cite{moreira2012inbreast} contains 115 exams from Centro Hospitalar de S. Jo\~{a}o in Portugal.
    Following~\cite{stadnick2021metarepository}, we use the official testing set that contains 31 biopsy confirmed exams, with 15 malignant exams and 16 non-malignant exams.
    Unlike the meta-repository~\cite{stadnick2021metarepository}, we did not evaluate on DDSM~\cite{web:ddsm} since it is a relatively old dataset (released in 1997) with low image quality (images are digitised from X-ray films), and we could not evaluate on OPTIMAM~\cite{halling2020optimam}, CSAW-CC~\cite{dembrower2020multi} and NYU~\cite{wu2019dataset} datasets since the first two have restricted access (only partnering academic groups can gain partial access to the data~\cite{stadnick2021metarepository}), and the NYU dataset~\cite{wu2019dataset} is private.

    \subsection{Implementation Details}
    We pre-process each image to remove text annotations and background noise outside the breast region, then we crop and pad the pre-processed images to fit into the target image size $1536 \times 768$ to avoid distortion. During data loading process, we resize the input image pairs to 1536 x 768 pixels and flip all images so that the breast points toward the right hand size of the image. To improve training generalisation, we use data augmentation based on random vertical flipping and random affine transformation. 
    BRAIxMVCCL uses EfficientNet-b0~\cite{tan2019efficientnet}, initialized with ImageNet~\cite{russakovsky2015imagenet} pre-trained weights, as the backbone feature extractor. Our training relies on Adam optimiser~\cite{kingma2014adam} using a learning rate of 0.0001, weight decay of $10^{-6}$, batch size of 8 images and 20 epochs. We use ReduceLROnPlateau to dynamically control the learning rate reduction based on the BCE loss during model validation, where the reduction factor is set to 0.1. 
    Hyper-parameters are estimated using the validation set. 
    All experiments are implemented with Pytorch~\cite{paszke2019pytorch} and conducted on an NVIDIA RTX 3090 GPU (24GB). The training takes 23.5 hours on ADMANI-1. The testing takes 42.17ms per image.
    For the fine-tuning of GMIC~\cite{shen2021interpretable} on ADMANI-1,2 datasets, we load the model checkpoint (trained on around 0.8M images from NYU training set) from the
    official github repository\footnote{\url{https://github.com/nyukat/GMIC}} and follow hyper-parameter setting from the author's github\footnote{\url{https://github.com/nyukat/GMIC/issues/4}}. 
    For fine-tuning DMV-CNN, we load the model checkpoint from the official  repository\footnote{\url{https://github.com/nyukat/breast_cancer_classifier}} and fine-tune the model with using the Adam optimiser~\cite{kingma2014adam} with learning rate of $10^{-6}$, a minibatch of size 8 and weight decay of $10^{-6}$. 
    The classification results are assessed with the area under the receiver operating characteristic curve (AUC-ROC) and area under the precision-recall curve (AUC-PR).

    \subsection{Results}
    Table~\ref{tab:admani} shows the testing AUC-ROC results on ADMANI-1,2.
    We calculate the \textbf{image-level} bootstrap AUC with 2,000 bootstrap replicates, from which we present the mean
    and the lower and upper bounds of the 95\% confidence interval (CI).
    Our model shows significantly better results than the SOTA model GMIC~\cite{shen2021interpretable} with a 5.89\% mean improvement.
    We evaluate the generalisation of the models trained on ADMANI-1 
    on the large-scale dataset ADMANI-2. 
    It is worth mentioning that the ADMANI-2 dataset contains images produced by machines not used in training (i.e., GE
    MEDICAL SYSTEMS, Philips Medical Systems and Agfa) and by a different image processing algorithm for the SIEMENS machine. 
    On ADMANI-2, our model achieves a mean AUC-ROC of 92.60\%, which is 9.52\% better than the baseline EfficientNet-b0~\cite{tan2019efficientnet}, 15.47\% better than DMV-CNN and 8.21\% comparing with GMIC. According to the CI values, our results are significantly better than all other approaches.

\begin{figure}[t]
        \includegraphics[width=0.7\textwidth]{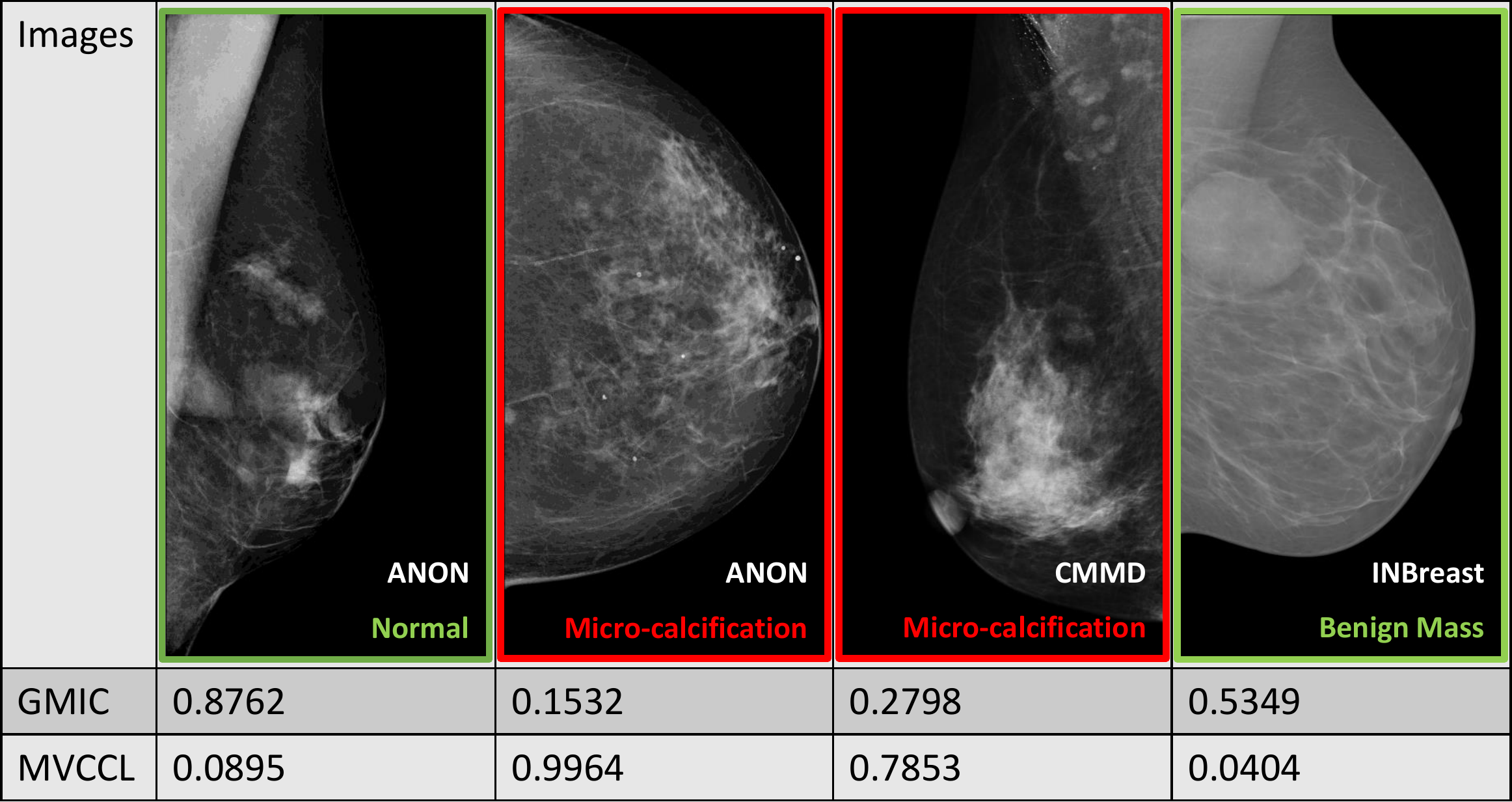}
        \centering
        \caption{Qualitative visual comparison of the classification of challenging cancer (\textcolor{red}{red} border) and non-cancer (\textcolor{green}{green} border) cases produced by GMIC and our BRAIxMVCCL on ADMANI, CMMD and INBreast datasets. The numbers below each image indicate the cancer probability prediction by each model.
        }
        \label{fig:visual}
    \end{figure}

    Table~\ref{tab:model_performance} shows the AUC-ROC and AUC-PR results of our model trained on ADMANI-1 on two publicly available datasets, namely INBreast~\cite{moreira2012inbreast} and CMMD~\cite{web:cmmd}, following the evaluation protocol in~\cite{stadnick2021metarepository}. 
    We report 95\% bootstrap confidence intervals with 2,000 bootstrap replicates based on \textbf{breast-level} evaluation\footnote{The breast-level result is obtained by averaging the predictions from both views.}.
    Our BRAIxMVCCL (single) achieves better results than other approaches, surpassing the mean AUC-ROC result by GMIC (single) by 6.8\% on INBreast and 2.7\% on CMMD, and also outperforming GMIC (top-5 ensemble) by 1.4\% on INBreast and 2.1\% on CMMD, even though our BRAIxMVCCL is not an ensemble method. 
    We show a qualitative visual comparison between GMIC~\cite{shen2021interpretable} and our BRAIxMVCCL in Fig.~\ref{fig:visual} where results demonstrates that our model can correctly classify challenging cancer and non-cancer cases missed by GMIC.

    \begin{table}[t]
    \begin{minipage}[t]{0.48\linewidth}
    \centering
    \caption{Image-level mean and 95\% CI AUC-ROC results for ADMANI datasets.}
        \resizebox{0.8\columnwidth}{!}{%
        \begin{tabular}{c|cc}
        \toprule
            Methods	                                    &	\makecell{ADMANI-1\\(test set)}	                &	\makecell{ADMANI-2\\(whole set)} \\ \midrule\midrule
            EfficientNet-b0~\cite{tan2019efficientnet}	&   \makecell{0.886\\(0.873-0.890)} 	&	\makecell{0.838\\(0.828-0.847)} \\ \hline
            DMV-CNN~\cite{shen2019deep}	                & \makecell{0.799\\(0.780-0.817)}	    &	\makecell{0.771\\(0.766-0.777} \\ \hline
            GMIC~\cite{shen2021interpretable}	        & \makecell{0.889\\(0.870-0.903)}	&	\makecell{0.844\\(0.835-0.853)} \\ \hline
            BRAIxMVCCL	                                    &	\textbf{\makecell{0.948\\(0.937-0.953)}}   &	\textbf{\makecell{0.926\\(0.922-0.930)}}
            \\ \hline\bottomrule
        \end{tabular}
        }
    \label{tab:admani}
    \end{minipage}
    \hspace{0.04cm}
    \begin{minipage}[t]{0.48\linewidth}
        \centering
        \caption{Ablation study of key components of our BRAIxMVCCL model.}
        \resizebox{0.8\columnwidth}{!}{%
        \begin{tabular}{ c|c|c|c|c}
        \toprule
            Fusion 	    & SA               &	LCM	         	&	GCM	            &	Testing AUC \\\midrule\midrule
            \checkmark	&		           &	                &	                &	0.9088   \\
            \checkmark	&	\checkmark     &			        &		            &	0.9183	\\
            \checkmark	&	               &	\checkmark	    &		            &	0.9368	\\
            \checkmark	&	               &		            &	\checkmark	    &	0.9241	\\
            \checkmark	&	\checkmark     &		            &	\checkmark	    &	0.9353	\\
            \checkmark	&                  &	\checkmark	    &	\checkmark	    &	\textbf{0.9478}	\\ \hline\bottomrule
        \end{tabular}}
        
    \label{tab:ablation}
    \end{minipage}
    \end{table}
    
    \input{table_model_performance}

    \subsection{Ablation Study}
    To evaluate the effectiveness of each component of our model, we show an ablation study based on testing AUC-ROC results in Table~\ref{tab:ablation}. 
    The study starts with the backbone model EfficientNet-b0~\cite{tan2019efficientnet}, which achieves 88.61\% (shown in first row of Tab.~\ref{tab:admani}). 
    Taking the representations extracted by the backbone model obtained from the two views and fusing them by concatenating the representations, and passing them through the classification layer (column 'Fusion' in Tab.~\ref{tab:ablation}) increases 
    the performance to 90.88\% (first row in Tab.~\ref{tab:ablation}). 
    To demonstrate the effectiveness of our local co-occurrence module (LCM),
    we compare it against a self-attention (SA) module~\cite{wang2018non, vaswani2017attention}, which integrates cross-view features and performs the self-attention mechanism. 
    We show that our proposed LCM is 1.97\% better than SA. 
    The model achieves 92.41\% with our global consistency module (GCM) and can be further improved to 93.53\% by using a self-attention (SA) module~\cite{wang2018non, vaswani2017attention} to combine the cross-view global features. 
    Finally, replacing SA by LCM further boosts the result by 1.25\%. Our proposed model achieves a testing AUC of 94.78\% which improves the baseline model by 6.17\%. 

\section{Conclusion}

We proposed a novel multi-view system for the task of high-resolution mammogram classification, which analyses global and local information of unregistered ipsilateral views. 
We developed a global consistency module to model the consistency between global feature representations and a local co-occurrence module to model the region-level correspondence in a weakly supervised scenario. 
We performed extensive experiments on large-scale clinical datasets and also evaluated the model on cross-domain datasets. We achieved SOTA classification performance and generalisability across datasets. In future work, we plan to study partial multi-view learning to resolve the issue of dependence on multi-view data.

\subsubsection{Acknowledgement.}
This work is supported by funding from the Australian Government under the Medical Research Future Fund - Grant MRFAI000090 for the Transforming Breast Cancer Screening with Artificial Intelligence (BRAIx) Project. We thank the St Vincent’s Institute of Medical Research for providing the GPUs to support the numerical calculations in this paper.


%

\bibliographystyle{splncs04}
\bibliography{bibliography}

\end{document}

%% file: table_model_performance.tex
\begin{table*}[t]
    \caption{Testing results show breast-level estimates of AUC-ROC and AUC-RP on INBreast~\cite{moreira2012inbreast} and CCMD~\cite{web:cmmd}. Results are reported with 95\% CI calculated from bootstrapping with 2,000 replicates. 
    Results from other methods are obtained from~\cite{stadnick2021metarepository}.
    }
    \centering
    \resizebox{1\textwidth}{!}{
    \begin{tabular}{@{}cccccccccc@{}}
        \toprule
        &&\multicolumn{8}{c}{\textbf{Models}}\\ 
        \cmidrule(l){3-10}
        
        \multirow{2}{*}[1.7em]{\textbf{Dataset}}
        & \multirow{2}{*}[1.7em]{\textbf{AUC}}
        & \makecell{\textit{End2end}\cite{shen2017end}\\(\textit{DDSM})} 
        & \makecell{\textit{End2end}\cite{shen2017end}\\(\textit{INbreast})} 
        & \makecell{\textit{Faster}\cite{ren2015faster}\\\textit{R-CNN}}
        & \textit{DMV-CNN}\cite{shen2019deep} 
        & \makecell{\textit{GMIC}\cite{shen2021interpretable}\\(\textit{single})} 
        & \makecell{\blue{\textbf{\textit{GMIC}}}\cite{shen2021interpretable}\\\blue{\textbf{(\textit{top-5 ensemble})}}}
        & \textit{GLAM}\cite{liu2021weakly}
        & \red{\textbf{\makecell{\textit{BRAIxMVCCL}\\(\textit{single})}}}  \\
        \midrule
        
         
        
        
        \multirow{2}{*}{INBreast}
        & ROC 
        & \makecell{0.676\\(0.469-0.853)} 
        & \makecell{0.977\\(0.931-1.000)}
        & \makecell{0.969\\(0.917-1.000)}
        & \makecell{0.802\\(0.648-0.934)} 
        & \makecell{0.926\\(0.806-1.000)} 
        & \blue{\textbf{\makecell{0.980\\(0.940-1.000)}}}
        & \makecell{0.612\\(0.425-0.789)}
        & \red{\textbf{\makecell{0.994\\(0.985-1.000)}}} \\
        
        & PR  
        & \makecell{0.605\\(0.339-0.806)} 
        & \makecell{0.955\\(0.853-1.000)}
        & \makecell{0.936\\(0.814-1.000)}
        & \makecell{0.739\\(0.506-0.906)} 
        & \makecell{0.899\\(0.726-1.000)} 
        & \blue{\textbf{\makecell{0.957\\(0.856-1.000)}}}
        & \makecell{0.531\\(0.278-0.738)} 
        & \red{\textbf{\makecell{0.986\\(0.966-1.000)}}} \\
        \midrule
        
        
        
        \multirow{2}{*}{CMMD$^\dagger$}
        & ROC 
        & \makecell{0.534\\(0.512-0.557)} 
        & \makecell{0.449\\(0.428-0.473)}
        & \makecell{0.806\\(0.789-0.823)} 
        & \makecell{0.740\\(0.720-0.759)} 
        & \makecell{0.825\\(0.809-0.841)}
        & \blue{\textbf{\makecell{0.831\\(0.815-0.846)}}}
        & \makecell{0.785\\(0.767-0.803)} 
        & \red{\textbf{\makecell{0.852\\(0.840-0.863)}}} \\
        
        & PR 
        & \makecell{0.517\\(0.491-0.544)} 
        & \makecell{0.462\\(0.438-0.488)}
        & \makecell{0.844\\(0.828-0.859)} 
        & \makecell{0.785\\(0.764-0.806)} 
        & \makecell{0.854\\(0.836-0.869)}
        & \blue{\textbf{\makecell{0.859\\(0.842-0.875)}}}
        & \makecell{0.818\\(0.798-0.837)} 
        & \red{\textbf{\makecell{0.876\\(0.864-0.887)}}} \\
        
        
        \bottomrule
    \end{tabular}
    }
    \label{tab:model_performance}
\end{table*}